\documentclass[11pt]{article}

\usepackage[preprint]{acl}

\usepackage{times}
\usepackage{latexsym}
\usepackage[T1]{fontenc}
\usepackage[utf8]{inputenc}
\usepackage{microtype}
\usepackage{inconsolata}

\usepackage{amsmath,amssymb,amsfonts}
\usepackage{tikz}
\usetikzlibrary{arrows.meta, decorations.pathreplacing}
\usepackage{graphicx}
\usepackage{booktabs}
\usepackage{array}
\usepackage{multirow}
\usepackage{arydshln}

\title{Knowledge-Preserving LLM Architecture Standardization:\\
Analytic Normalizer Mapping with Equal-Size Distillation}

\author{
    \textbf{Kisu Yang}\textsuperscript{1,2} \quad
    \textbf{Yoonna Jang}\textsuperscript{3} \quad
    \textbf{Hyeonseok Moon}\textsuperscript{4} \quad
    \textbf{Hwanseok Jang}\textsuperscript{5} \quad
    \textbf{Taewoo Lee}\textsuperscript{1} \\
    \textbf{Hyungjin Lee}\textsuperscript{1} \quad
    \textbf{Jeseung Lee}\textsuperscript{1} \quad
    \textbf{Juhyoung Park}\textsuperscript{6} \quad
    \textbf{Heuiseok Lim}\textsuperscript{2}\thanks{Corresponding author.} \\
    \textsuperscript{1}VAIV Company \quad
    \textsuperscript{2}Korea University \quad
    \textsuperscript{3}Hanwha Aerospace \\
    \textsuperscript{4}Sookmyung Women's University \quad
    \textsuperscript{5}SEMES \quad
    \textsuperscript{6}DAS \\
}

\begin{document}
\maketitle

\begin{abstract}
Competent pretrained LLMs are often locked into architectures that mainstream tooling no longer supports, which strands their knowledge. The conventional remedy, \textit{uptraining}, needs heavy compute plus a corpus covering everything the model knows. We study a common class of such conversions, \textit{architecture standardization}, converting a model's normalization layers and interface to the Llama-family template while preserving dimensions, tokenizer, and attention layout. Our \textbf{KEPT} combines direct parameter copying, a training-free LayerNorm-to-RMSNorm initialization (OPM) that is near-optimal under activation assumptions we verify on three models, and standard equal-size distillation from the frozen source, without access to the source model's pretraining corpus. Converted from Orion-14B with $\sim 123$M tokens in four A100-days, Llamion stays within one point of the source's H6 average and MT-Bench score, shows no significant difference on Korean MMLU, HumanEval, or MBPP, and enters a Korean leaderboard closed to the custom-loader source, where its 66.87\% KoMMLU accuracy exceeds the next-best similar-size entry by more than seven points at submission time. In controlled runs, the default normalizer initialization fails to recover the model under the same budget, and both FLOPs-matched uptraining and sequence-level distillation trail on MT-Bench.
\end{abstract}

\section{Introduction}
\label{sec:introduction}

The Transformer~\citep{c:22} has become the universal template for large language models (LLMs), yet within it the open-source community makes divergent low-level choices, such as LayerNorm~\citep{ba2016layernormalization} vs.\ RMSNorm~\citep{zhang2019root}, multi-head attention (MHA) vs.\ grouped-query attention (GQA)~\citep{ainslie2023gqa}, absolute, ALiBi, or rotary (RoPE)~\citep{su2024roformer} position encodings, and GeLU vs.\ SwiGLU activations. The field has converged on a dominant \emph{family-level template}, a decoder-only stack with RoPE, RMSNorm, and SwiGLU shared by Llama-2/3~\citep{touvron2023llama,dubey2024llama} and Qwen2.5~\citep{qwen2024qwen25}, but earlier or third-party checkpoints remain valuable while increasingly out of step with it. Practitioners therefore regularly need to convert a competent pretrained model to the template, and not converting has a direct cost. Leaderboards and production stacks that enforce \texttt{trust\_remote\_code\,=\,False} cannot load a checkpoint that requires custom loading code, so its knowledge is left stranded.

Architectural transformation often causes knowledge loss~\citep{komatsuzaki2022sparse,mercat2024linearizing}. Even mechanical conversions such as MHA$\to$GQA alter the effective computational graph enough to partially destroy learned knowledge. The conventional remedy, uptraining (continued training so the modified network re-discovers lost capabilities), works but at considerable cost. The GQA recipe~\citep{ainslie2023gqa} uses ${\sim}5\%$ of original pretraining compute, and the new corpus must cover everything the model previously knew, since uncovered knowledge is subject to catastrophic forgetting~\citep{goodfellow2013empirical,kar2022preventing}. For modern LLMs whose pretraining corpora are not public, such curation is often impossible.

We propose a different remedy for a common class of such conversions. \textbf{KEPT}\footnote{The acronym slightly rearranges the word order of \emph{Efficient Knowledge Preservation for Transformation} for clarity.} does not retrain the modified network on a new corpus. It aligns the network's outputs with those of the original on a small generic corpus. The frozen original acts as an equal-size teacher, and the modified network, initialized by parameter-level mapping from the teacher, learns to reproduce its hidden states or logits. Because the alignment target is the teacher's behavior, KEPT needs no access to the source model's pretraining corpus and produces a drop-in replacement within standard Llama-template tooling. Keeping the student at the source's size is what makes this cheap. Every tensor outside the normalizers is copied verbatim, whereas compressive distillation into a smaller student has needed far more data, for instance about 94B tokens to distill an 8B model into a 4B one~\citep{sreenivas2024minitron}, while still giving up capacity. KEPT comprises \emph{Normal Parameter Mapping} (NPM) for unchanged modules, \emph{Optimized Parameter Mapping} (OPM) for changed modules admitting a principled initialization, and \emph{Cross-architecture Knowledge Distillation} (XKD) for residual alignment (Figure~\ref{fig:knowledge_transfer}).

\begin{figure}[t]
  \centering
  \resizebox{0.96\linewidth}{!}{
\begin{tikzpicture}[
  font=\scriptsize,
  mod/.style={draw, rounded corners=1pt, minimum width=2.5cm, minimum height=0.4cm, inner sep=1pt, align=center},
  srcmod/.style={mod, fill=blue!8, draw=blue!40!black!60},
  tgtmod/.style={mod, fill=green!9, draw=green!45!black!60},
  srcnorm/.style={mod, fill=orange!22, draw=orange!70!black},
  tgtnorm/.style={mod, fill=orange!22, draw=orange!70!black},
  npm/.style={-{Stealth[length=1.6mm]}, thick, black!70},
  opm/.style={-{Stealth[length=2mm]}, very thick, orange!85!black},
  lbl/.style={font=\tiny, inner sep=0.5pt},
]
\def\dy{0.47}
\def\rx{4.05}
\node[srcmod]  (l0) at (0,0)        {Token Emb.\ $V{\times}d$};
\node[srcnorm] (l1) at (0,\dy)      {\textbf{LayerNorm} $(\gamma,\beta)$};
\node[srcmod]  (l2) at (0,2*\dy)    {MHA (RoPE), 40 heads};
\node[srcnorm] (l3) at (0,3*\dy)    {\textbf{LayerNorm} $(\gamma,\beta)$};
\node[srcmod]  (l4) at (0,4*\dy)    {SwiGLU MLP};
\node[srcnorm] (l5) at (0,5*\dy)    {\textbf{Final LN} $(\gamma,\beta)$};
\node[srcmod]  (l6) at (0,6*\dy)    {LM Head};
\node[tgtmod]  (r0) at (\rx,0)      {Token Emb.\ $V{\times}d$};
\node[tgtnorm] (r1) at (\rx,\dy)    {\textbf{RMSNorm} $(\theta)$};
\node[tgtmod]  (r2) at (\rx,2*\dy)  {MHA (RoPE), 40 heads};
\node[tgtnorm] (r3) at (\rx,3*\dy)  {\textbf{RMSNorm} $(\theta)$};
\node[tgtmod]  (r4) at (\rx,4*\dy)  {SwiGLU MLP};
\node[tgtnorm] (r5) at (\rx,5*\dy)  {\textbf{Final RMSNorm} $(\theta)$};
\node[tgtmod]  (r6) at (\rx,6*\dy)  {LM Head};
\node[align=center, font=\scriptsize\bfseries] at (0,6*\dy+0.58) {Orion (frozen source)};
\node[align=center, lbl, text=black!60] at (0,6*\dy+0.35) {stranded: \texttt{trust\_remote\_code=True}};
\node[align=center, font=\scriptsize\bfseries] at (\rx,6*\dy+0.58) {Llamion (trained student)};
\node[align=center, lbl, text=black!60] at (\rx,6*\dy+0.35) {loads with \texttt{trust\_remote\_code=False}};
\foreach \i in {0,2,4,6} { \draw[npm] (l\i.east) -- (r\i.west); }
\foreach \i in {1,3,5}   { \draw[opm] (l\i.east) -- (r\i.west); }
\node[lbl, text=black!55, fill=white, inner sep=0.3pt] at (0.5*\rx, 0.115) {unchanged};
\draw[decorate, decoration={brace, amplitude=3pt}, black!60]
  (-1.38,0.56*\dy) -- (-1.38,4.44*\dy) node[midway, left=3pt, lbl, rotate=90, anchor=south] {$\times 40$ layers};
\draw[npm] (-1.15,-0.46) -- (-0.75,-0.46) node[right, lbl, text=black!75] {\textbf{NPM}: copy verbatim};
\draw[opm] (1.2,-0.46) -- (1.6,-0.46) node[right, lbl, text=orange!75!black] {\textbf{OPM}: $\theta \leftarrow \gamma$, discarded $\beta \Rightarrow \|\beta\|_2^2$ residual};
\draw[-{Stealth[length=2mm]}, very thick, dashed, teal!70!black]
  (0,-0.92) -- node[below=0.5mm, lbl, text=teal!45!black, align=center]
  {Equal-size distillation (MSE on hidden states or logits)}
  (\rx,-0.92);
\end{tikzpicture}}
  \caption{Overview of KEPT. Unchanged modules are copied verbatim (NPM), and each LayerNorm becomes an RMSNorm initialized analytically at $\theta = \gamma$ (OPM, L2-optimal under Appendix~\ref{sec:appendix_proof}'s assumptions), leaving the discarded bias $\beta$ as the residual, which equal-size distillation from the frozen source (XKD) absorbs.}
  \label{fig:knowledge_transfer}
\end{figure}

Our main theoretical contribution is a closed-form result for OPM. The RMSNorm weights that minimize the expected L2 output-alignment error against a replaced LayerNorm equal the LayerNorm weights themselves, under near-zero-mean, on-average-symmetric activation assumptions motivated by weight-decay training (Appendix~\ref{sec:appendix_proof}). Measurements on Orion-14B, Falcon-7B, and Pythia-12B confirm that these assumptions hold to within a few percent. OPM is therefore training-free, and the irreducible residual, the discarded LayerNorm bias, is what XKD absorbs.

\paragraph{Scope.} The empirical scope is narrow. The case study replaces the normalizers (LayerNorm$\to$RMSNorm) and standardizes the interface and prompt template, while hidden dimensions, layer count, attention layout, activation, and tokenizer are preserved. Harder conversions such as MHA$\to$GQA, GeLU$\to$SwiGLU, and cross-width or cross-tokenizer transfer are not demonstrated. For the first two we give an analytic account of when closed-form mappings exist (Appendix~\ref{sec:appendix_opm_extensions}).

We instantiate KEPT on Orion-14B~\citep{chen2024orion}, a model competent in East Asian languages but built on a non-standard stack (LayerNorm, a loader-level attention implementation, and custom code requiring \texttt{trust\_remote\_code}), and convert it into the Llama-family template~\citep{touvron2023llama}. The result, Llamion, is directly usable in mainstream tooling (Section~\ref{sec:why_llama}). With ${\sim}123$M training tokens on one A100 over four days, Llamion stays within one point of Orion on H6 and MT-Bench and shows no significant difference on Korean MMLU, HumanEval, or MBPP. It also attains 66.87\% KoMMLU on the Open Ko LLM Leaderboard, which its source cannot enter, more than seven points above the next-best entry in its parameter class at submission time. Capabilities absent from the transfer corpus, namely multi-turn dialogue, code generation, and long-context use, survive the conversion (Section~\ref{sec:analysis}).

\paragraph{Contributions.}
(1) \textbf{KEPT}, a low-compute pipeline (NPM, OPM, XKD) for knowledge-preserving architecture standardization of pretrained LLMs without access to the source model's pretraining corpus.
(2) An analytic result (Appendix~\ref{sec:appendix_proof}) that, with all copied tensors held byte-identical, the L2-optimal RMSNorm weights replacing a LayerNorm equal the LayerNorm weights up to a negligible $O(1/n)$ correction, under activation assumptions that we verify on three LayerNorm-based LLMs (Section~\ref{sec:optimized_parameter_mapping}). The result yields a training-free initialization whose irreducible residual is the discarded bias.
(3) The released Llamion checkpoints (Base, Chat, LongChat), together with controlled comparisons (Section~\ref{sec:exp_controlled}) showing that the default normalizer initialization does not recover the model under the same budget and that XKD stays ahead of FLOPs-matched uptraining and sequence-level distillation on MT-Bench.
(4) An analytic account of when closed-form mappings exist beyond LayerNorm$\to$RMSNorm (Appendix~\ref{sec:appendix_opm_extensions}), with an optimality derivation for GQA mean-pooling at the level of a projection-alignment surrogate and an impossibility argument for unit-level GeLU$\to$SwiGLU.

\section{Related Work}
\label{sec:related_work}

\paragraph{Uptraining and architectural transformation.}
Uptraining, the continued training of an architecturally modified network~\citep{komatsuzaki2022sparse,mercat2024linearizing}, is the dominant remedy when an LLM's architecture is altered. GQA~\citep{ainslie2023gqa} reports recovery at ${\sim}5\%$ of pretraining compute. Linearizing LLMs~\citep{mercat2024linearizing} and Mamba-in-Llama~\citep{wang2024mambainllama} apply the principle to more aggressive transitions, the latter distilling a Transformer into a Mamba\textendash attention hybrid reusing attention weights. A complementary line pursues cheap modification via principled initialization plus light continued training. Sparse Upcycling~\citep{komatsuzaki2022sparse} expands MLPs into MoE, SOLAR~\citep{kim2023solar} up-scales depth, LLaMA-Pro~\citep{wu2024llamapro} inserts identity-initialized blocks, and EEVE~\citep{kim2024efficient} expands vocabulary for Korean. Closest to our normalizer mapping, \citet{jiang2023prermsnorm} show Pre-LN Transformers are exactly convertible to Pre-RMSNorm by editing adjacent parameters, as used by SliceGPT~\citep{ashkboos2024slicegpt}. Section~\ref{sec:optimized_parameter_mapping} details how OPM relates under the bias-free template.

\paragraph{Knowledge distillation.}
Classical knowledge distillation (KD) trains a small student to mimic a larger teacher's outputs~\citep{bucilua2006model,ba2014deep,hinton2015distilling}, classified as offline, online, or self-distillation~\citep{hinton2015distilling,zhang2018deep,zelikman2022star,wang2022self,xu2023baize}, with surveys in \citet{xu2024survey,gou2021knowledge}. KEPT's XKD step is structurally an equal-size offline distillation~\citep{hinton2015distilling}, for whose objective we claim no novelty, and this sizing distinction matters. Classical KD compresses and thereby incurs capacity-induced loss of parametric knowledge no matter how well the alignment is performed, while XKD does not because student and teacher have the same capacity. \citet{zhong2023seeking} is closest in spirit, extracting parametric knowledge for transfer, but only for specific domains. Recent cross-model transfer methods likewise move a single capability onto an already pretrained target, such as a task adapter~\citep{jung2026titok} or a capability direction in activation space~\citep{balasubramanian2026masterkey}, whereas KEPT carries a model's entire pretrained knowledge into a new architecture.

\section{Methodology}
\label{sec:methodology}

KEPT preserves the knowledge of a pretrained source model $M_{src}$ when its architecture changes to a target architecture $A_{tgt}$, instantiating $A_{tgt}$ as a target model $M_{tgt}$ in two stages, (i) a parameter-level initialization, with NPM (Section~\ref{sec:normal_parameter_mapping}) for unchanged modules and OPM (Section~\ref{sec:optimized_parameter_mapping}) for changed modules, and (ii) XKD (Section~\ref{sec:knowledge_transfer}), which closes the residual mismatch by aligning $M_{tgt}$'s outputs with those of the frozen $M_{src}$. The transformation is one-directional. $M_{src}$ is discarded after training, and no feature fusion or network merging is involved.

\begin{table*}[!t]
\centering
\small
\setlength{\tabcolsep}{4pt}
\begin{tabular}{@{}clccccccc@{}}
\toprule
Model & Method & ARC & HellaSwag & MMLU & TruthfulQA & Winogrande & GSM8K & Average \\
\midrule
Base & --\,(original)             & 54.27             & 80.98    & 69.39 & 40.06 & 74.35    & 45.94    & 60.83    \\
Base & Map                        & 26.02             & 30.50             & 26.98             & 45.35    & 56.75             & \;\,1.52          & 31.19             \\
Base & Map\,+\,UT                 & 52.61             & 78.35             & 67.57             & 39.11             & 72.39             & 36.58             & 57.77 \\
\addlinespace[2pt]\hdashline\addlinespace[3pt]
Base & Map\,+\,XKD$_{h}$ (ours)   & 54.42 & 80.00             & 68.36             & 39.33             & 73.37             & 44.12             & 59.93 \\
Base & Map\,+\,XKD$_{z}$ (ours)   & \textbf{55.12}    & \textbf{80.40} & \textbf{69.40}    & \textbf{39.74}             & \textbf{73.77} & \textbf{44.70} & \textbf{60.53} \\
\midrule
Chat & --\,(original)             & 54.35    & 79.03    & 62.68    & 44.20 & 72.77 & 41.32 & 59.06    \\
Chat & Map                        & 25.17             & 34.25             & 31.17             & 51.21    & 60.38             & \;\,3.26          & 34.24             \\
Chat & Map\,+\,UT                 & 52.22             & 76.73             & 60.94             & 42.20             & 71.19             & 33.59             & 56.15 \\
\addlinespace[2pt]\hdashline\addlinespace[3pt]
Chat & Map\,+\,XKD$_{h}$ (ours)   & 54.01             & 78.31             & 61.51             & 42.39             & 72.22             & 40.41             & 58.14 \\
Chat & Map\,+\,XKD$_{z}$ (ours)   & \textbf{54.27} & \textbf{78.96} & \textbf{62.15} & \textbf{42.90}             & \textbf{73.01}    & \textbf{41.85}    & \textbf{58.86} \\
\bottomrule
\end{tabular}
\caption{H6 few-shot performance. Within each model group, \textbf{bold} marks the best score among the trained conversions. The original model is the reference. Map is the parameter-mapped model (NPM+OPM, Sections~\ref{sec:normal_parameter_mapping} and~\ref{sec:optimized_parameter_mapping}) with no further training, UT adds uptraining, and XKD$_{h}$ and XKD$_{z}$ add Cross-architecture Knowledge Distillation with the hidden-state and logit objectives. Sub-point differences are within single-run noise (Section~\ref{sec:experimental_settings}).}
\label{tab:h6}
\end{table*}

\subsection{Choice of Case Study}
\label{sec:why_llama}

We adopt Orion-14B~\citep{chen2024orion} as $M_{src}$ and the Llama-family architecture~\citep{touvron2023llama} as $A_{tgt}$, producing Llamion. The target is the family-level template of Section~\ref{sec:introduction}, introduced by Llama-2 and inherited essentially unchanged by Llama-3 and Qwen2.5, not any single checkpoint. The pair is an instance of the standardization class this paper targets, since Orion uses LayerNorm, a loader-level attention implementation (computationally standard MHA), and custom code requiring \texttt{trust\_remote\_code\,=\,True} in Transformers~\citep{wolf2019huggingface}, yet shares the template's dimensions, SwiGLU activation, and RoPE. The structural change exercised here is the normalizer replacement plus interface standardization. Larger divergences are analyzed in Appendix~\ref{sec:appendix_opm_extensions} but not demonstrated, and KEPT itself requires only that $M_{src}$ and $A_{tgt}$ share a stacked-block, residual-stream structure and the same hidden dimension.

\subsection{Normal Parameter Mapping}
\label{sec:normal_parameter_mapping}

NPM places the parameters of $M_{src}$ into the corresponding locations of $A_{tgt}$ for modules whose computation is unchanged, so the mapping is identity in the parameter tensor. The Llamion configuration matches Orion's dimensions, with $V = 84{,}608$, $d = 5{,}120$, MLP intermediate $15{,}360$, 40 attention heads of dim 128, and 40 decoder layers, with full multi-head attention preserved. Token embeddings, the LM head, and all attention/MLP projections are copied verbatim. The tokenizer retains Orion's subword model unchanged, so every embedding row keeps its meaning. Two string-level adaptations make the converted model compose with Llama-3-style tooling. The chat template is replaced with the Llama-3 format, and the Llama-3 special-token surface forms are assigned to existing reserved token IDs (Appendix~\ref{sec:appendix_reproducibility}). Neither changes any model parameter. We convert three Orion variants (Base, Chat, LongChat), yielding the corresponding Llamion variants. Per-module mapping details are summarized in Table~\ref{tab:module_mapping} (Appendix~\ref{sec:module_mapping}).

\subsection{Optimized Parameter Mapping}
\label{sec:optimized_parameter_mapping}

OPM handles modules whose computation has changed but for which a principled, training-free initialization exists. In Orion$\to$Llama, the only such module is the normalizer. $M_{src}$ uses LayerNorm~\citep{ba2016layernormalization} with weight $\gamma$ and bias $\beta$, while $A_{tgt}$ uses RMSNorm~\citep{zhang2019root} with weight $\theta$. Let $a = [a_{1}, \ldots, a_{n}]$ be the pre-norm activation with mean $\mu = \tfrac{1}{n}\sum_i a_i$, standard deviation $\sigma$, and root mean square $\mathrm{RMS}(a) = \sqrt{\tfrac{1}{n}\sum_i a_i^2}$:
\[
\mathrm{LN}_i = \frac{a_i - \mu}{\sigma}\, \gamma_i + \beta_i, \quad
\mathrm{RN}_i = \frac{a_i}{\mathrm{RMS}(a)}\, \theta_i.
\]
When $\mu \approx 0$, LayerNorm and RMSNorm coincide up to $\beta_i$. This regime is expected under AdamW training with weight decay (Orion's setting), and we verify it below. We therefore set $\theta_i \leftarrow \gamma_i$ and discard $\beta_i$. Appendix~\ref{sec:appendix_proof} provides the formal statement that, under expected-L2 output alignment with near-zero-mean, on-average-symmetric activations, the optimal RMSNorm weights equal the LayerNorm weights (up to a negligible $1/n$ width correction) independently of $\beta$, and the irreducible alignment error in the zero-mean limit is $\|\beta\|_2^2$, the discarded bias, which the subsequent XKD stage absorbs.

\paragraph{Empirical check of the assumptions.} We measured pre-norm activations at all 81 normalizer sites of Orion on held-out text in the four transfer languages (196,608 tokens) and report medians over sites. The near-zero-mean regime holds, with the 99th percentile of $|\mu|/\sigma$ at 0.010. The least-squares optimal RMSNorm weights computed from the measured moments deviate from $\gamma$ by 1.9\%. This exceeds the $1/n$ correction, so the symmetry assumption holds only approximately, but copying $\gamma$ remains close to optimal. The alignment error at $\theta = \gamma$ is 12.5 against $\|\beta\|_2^2 = 12.2$, so the residual is almost entirely the discarded bias. The stabilizing constant $\epsilon_0 = 10^{-5}$ that both normalizers add to their denominators, which the analysis omits, is negligible because the smallest observed $\mathrm{RMS}(a)^2$ exceeds it by a factor of $8.4{\times}10^{4}$. Falcon-7B~\citep{almazrouei2023falcon} and Pythia-12B~\citep{biderman2023pythia} give comparable values (Table~\ref{tab:act_stats}, Appendix~\ref{sec:appendix_proof}), so the regime is not specific to Orion. Training the normalizer weights alone for 100 steps lowers the KL divergence from the source from 4.03 to 0.27 when started at $\theta = \gamma$, but only from 38.72 to 35.28 when started at random (Appendix~\ref{sec:appendix_proof}). Estimating the weights from activation statistics brings no benefit in practice, since the least-squares weights start further from the source than $\theta = \gamma$ does (Section~\ref{sec:exp_controlled}).

\paragraph{Relation to exact normalization conversion.} That LayerNorm and RMSNorm coincide on zero-mean input is well known. Our contribution is its formalization as an optimality statement with a quantified residual. Pre-LN and Pre-RMSNorm Transformers can be made exactly equivalent by recentering the residual-stream writers and folding $\beta$ into subsequent linear-layer biases~\citep{jiang2023prermsnorm,ashkboos2024slicegpt}. We instead target the bias-free template default, in which copied tensors stay byte-identical. The stock LM head is bias-free, so the final norm's $\beta$ cannot be folded within the standard model class, which gives our $\|\beta\|_2^2$ residual. Section~\ref{sec:exp_controlled} evaluates the recentering half of the exact route, which modifies copied tensors, alongside OPM.

\subsection{Cross-architecture Knowledge Distillation}
\label{sec:knowledge_transfer}

After NPM and OPM, $M_{tgt}$ produces outputs close to but not identical to those of $M_{src}$. XKD closes the residual gap (Figure~\ref{fig:knowledge_transfer}) under one of two objectives. The name marks the setting (equal-size frozen-teacher alignment across an architectural conversion), not a new method, as both objectives are standard MSE alignment losses~\citep{hinton2015distilling,gou2021knowledge}. The hidden-state objective matches the per-layer residual stream:
\begin{equation}
\label{eq:loss_hidden}
\mathcal{L}_{h} = \frac{1}{L(N{+}1)} \sum_{j=0}^{L-1} \sum_{i=0}^{N} \big\| h_{i,j} - \hat{h}_{i,j} \big\|_{2}^{2},
\end{equation}
where $L$ is sequence length, $N$ is the number of decoder layers, and $h_{i,j}$/$\hat{h}_{i,j}$ are residual-stream activations from $M_{src}$/$M_{tgt}$ after layer $i$ at position $j$ (the $N{+}1$ slots include the embedding output). The logit objective matches the LM-head outputs $z_j$ and $\hat{z}_j$ of $M_{src}$ and $M_{tgt}$ at position $j$:
\begin{equation}
\label{eq:loss_logits}
\mathcal{L}_{z} = \frac{1}{L} \sum_{j=0}^{L-1} \big\| z_{j} - \hat{z}_{j} \big\|_{2}^{2}.
\end{equation}
The hidden-state objective pins the entire residual-stream trajectory, while the logit objective is more permissive, letting the internal trajectory drift as long as the final output stays aligned. The two objectives produce students of comparable final quality (differences are sub-point and within our stated interpretation threshold, Section~\ref{sec:experimental_settings}). We default to the logit objective because it converges faster and touches less teacher state per step.

\section{Experiments}
\label{sec:experiments}

We evaluate KEPT on English few-shot accuracy (H6), English multi-turn dialogue (MT-Bench), and Korean few-shot accuracy (KoMMLU). A complementary fine-tuning evaluation on four Korean generation tasks appears in Appendix~\ref{sec:appendix_ft_korean}.

\subsection{Settings}
\label{sec:experimental_settings}

We use the MIRACL corpus~\citep{10.1162/tacl_a_00595} restricted to Korean/English/Chinese/Japanese subsets, balanced to ${\sim}123$M tokens, with sequences concatenated and packed to length 4{,}096~\citep{liu2019roberta}. Per-device batch size is 1 (the per-token logit is the effective sample), learning rate $1{\times}10^{-5}$ with cosine annealing, 100 warmup steps, and 30K total steps. Learning rates above $3{\times}10^{-5}$ produced instability. Training takes ${\sim}4$ days on one A100 80GB, with optimizer states offloaded to host memory (Appendix~\ref{sec:appendix_reproducibility}). We evaluate on (i) H6 (ARC~\citep{clark2018think}, HellaSwag~\citep{zellers2019hellaswag}, MMLU~\citep{hendrycks2020measuring}, TruthfulQA~\citep{lin2021truthfulqa}, Winogrande~\citep{sakaguchi2021winogrande}, GSM8K~\citep{cobbe2021training}) via LM Evaluation Harness~\citep{eval-harness} with Open LLM Leaderboard~\citep{open-llm-leaderboard} few-shot configurations, (ii) MT-Bench~\citep{zheng2024judging} judged by \texttt{gpt-4-0613} with single-answer grading via FastChat, and (iii) KoMMLU on the Open Ko LLM Leaderboard~\citep{park2024open}. Full hyperparameter and tokenizer-ID detail appears in Appendix~\ref{sec:appendix_reproducibility}.

\paragraph{Determinism and variance.} The numbers in Tables~\ref{tab:h6} and~\ref{tab:mt_bench} come from a single training run and a single evaluation pass, since a 4-day-per-run budget on one GPU made seed repetition impractical. The H6 evaluations are deterministic given the harness few-shot configuration (greedy log-likelihood scoring, fixed prompts), so run-to-run noise is confined to MT-Bench's LLM judge and training stochasticity. We accordingly interpret only gaps that are large relative to known variability, such as the seven-point KoMMLU margin against a between-checkpoint variance of one to two points, and never sub-point differences.

\paragraph{Controlled runs and agreement metrics.} Section~\ref{sec:exp_controlled} reports additional runs of 1K steps (4.1M tokens) on the same data stream, each a single training run. They were evaluated in a separate, later pass in which MT-Bench is judged three times by \texttt{gpt-4-0613}, and we report the mean and, where relevant, the standard deviation over the three passes. Absolute MT-Bench scores in this pass are higher than in Table~\ref{tab:mt_bench} (7.49 against 7.116 for the same source model), so these runs are compared with the source as re-evaluated in the same pass and never with Tables~\ref{tab:h6} and~\ref{tab:mt_bench}. We also measure agreement with the source directly on held-out text, using the KL divergence between the source's and the student's next-token distributions and the fraction of positions where their top-1 predictions coincide.

\subsection{H6 Benchmark and Ablation Analysis}
\label{sec:exp_h6}

Table~\ref{tab:h6} is an ablation isolating each KEPT component, comparing (a) the source model Orion as upper bound, (b) parameter mapping alone (NPM/OPM, no training), (c) mapping plus 30K steps of uptraining (UT), i.e.\ standard next-token cross-entropy on the same packed MIRACL stream, and (d, e) the full pipeline with the hidden-state ($\mathrm{XKD}_{h}$) or logit ($\mathrm{XKD}_{z}$) objective. UT and both XKD variants share data, optimizer settings, and step budget, and no separate hyperparameter sweep was run for UT beyond these shared settings.

\paragraph{Contribution of parameter mapping.} With zero training, the mapped student (Base) reaches 31.19 on H6, above chance level (${\sim}25$\% on most subtasks). This is the parameter-mapping floor, showing that NPM/OPM alone is a non-trivial but insufficient starting point. Its apparent TruthfulQA advantage over the source is a known artifact of degraded models on that task's format and is not interpreted.

\paragraph{Contribution of XKD on top of mapping.} The jump from (b) 31.19 to (e) 60.53 on Base is $+29.3$ points (Chat: 34.24$\to$58.86, $+24.6$), so almost the entire recovery is attributable to XKD. The mapping thus acts as an initialization that a small XKD budget (123M tokens) can complete, and Section~\ref{sec:exp_controlled} shows that the same kind of budget fails from the default normalizer initialization.

\paragraph{XKD versus uptraining at matched data and steps.} At identical data and step count, $\mathrm{XKD}_{z}$ beats UT by ${+}2.76$ on Base (60.53 vs.\ 57.77) and ${+}2.71$ on Chat. The gap is modest in H6's discriminative regime but much larger in the generative regime, where on MT-Bench (Table~\ref{tab:mt_bench}, scale 1--10) UT-Chat reaches 5.709 vs.\ $\mathrm{XKD}_{z}$-Chat 7.034 (${+}1.33$, a $23\%$ relative gain). The gap is thus larger on the generative metric. Matching data and steps does not match total computation, because each XKD step adds a forward pass through the frozen teacher, about $1.33\times$ the floating-point operations of an uptraining step (Appendix~\ref{sec:appendix_reproducibility}). Together with the untuned UT recipe, this means the result holds for the shared recipe and is not a claim at equal total computation. Section~\ref{sec:exp_controlled} repeats the comparison at matched FLOPs and at a second learning rate.

\begin{figure}[t]
  \centering
  \includegraphics[width=0.66\linewidth]{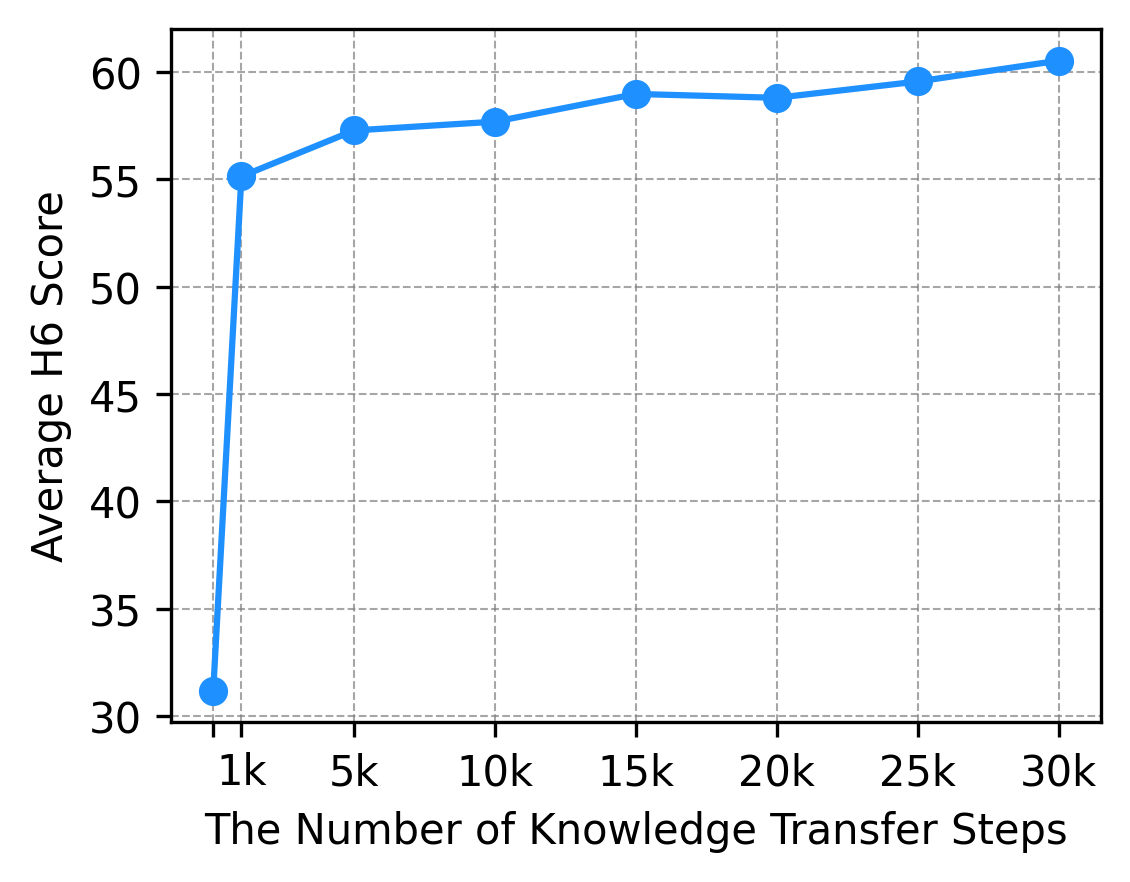}
  \caption{Average H6 score versus the number of XKD-logit training steps. The score rises sharply within the first 1K steps and continues to improve smoothly, consistent with the strong initialization from NPM/OPM.}
  \label{fig:avg_h6_scores}
\end{figure}

\paragraph{Hidden-state vs.\ logit objective.} The two objectives finish within noise of each other (59.93 vs.\ 60.53 on Base, 58.14 vs.\ 58.86 on Chat), a tie rather than a ranking per our variance policy (Section~\ref{sec:experimental_settings}). We default to the logit objective on practical grounds (faster convergence, and only final logits needed). Figure~\ref{fig:avg_h6_scores} shows scores still rising at 30K steps, so the residual ${\sim}0.3$-point gap from Orion is likely closable with more budget.

\subsection{Multi-turn Conversation (MT-Bench)}
\label{sec:exp_mtbench}

\begin{table}[t]
\centering
\small
\setlength{\tabcolsep}{3pt}
\begin{tabular}{@{}clccc@{}}
\toprule
Model & Method & Turn-1 & Turn-2 & Avg. \\
\midrule
Base & --\,(original)               & 6.481    & 4.613 & 5.547    \\
Base & Map                          & 1.000             & 1.000             & 1.000             \\
Base & Map\,+\,UT                   & 5.300             & 3.763             & 4.532 \\
\addlinespace[2pt]\hdashline\addlinespace[3pt]
Base & Map\,+\,XKD$_{h}$            & 6.250             & \textbf{4.656}    & 5.453 \\
Base & Map\,+\,XKD$_{z}$            & \textbf{6.456} & 4.525             & \textbf{5.491} \\
\midrule
Chat & --\,(original)               & 7.744    & 6.488    & 7.116    \\
Chat & Map                          & 1.338             & 1.113             & 1.226             \\
Chat & Map\,+\,UT                   & 6.556             & 4.861             & 5.709 \\
\addlinespace[2pt]\hdashline\addlinespace[3pt]
Chat & Map\,+\,XKD$_{h}$            & \textbf{7.744} & 6.013             & 6.879 \\
Chat & Map\,+\,XKD$_{z}$            & 7.731             & \textbf{6.338} & \textbf{7.034} \\
\bottomrule
\end{tabular}
\caption{MT-Bench results (1--10 scale, \texttt{gpt-4-0613} judge). Method labels as in Table~\ref{tab:h6}.}
\label{tab:mt_bench}
\end{table}

Table~\ref{tab:mt_bench} reports MT-Bench scores. KEPT preserves multi-turn dialogue quality within 0.1 points of Orion-Chat ($7.034$ vs.~$7.116$). Mapping-only collapses to the floor ($1.000$ Base, $1.226$ Chat), where its degraded text receives the judge's minimum score. The contrast with its above-chance H6 score has a simple explanation, since H6 is a discriminative top-1 task tolerant of a noisy log-likelihood landscape, while free-form generation is far more sensitive to next-token-distribution perturbations. This explains why the gap between UT and XKD is modest on H6 but large on MT-Bench, and it suggests that work on architectural transformation should report generative metrics alongside discriminative ones.

\subsection{Controlled Comparisons at a Fixed Budget}
\label{sec:exp_controlled}

\begin{table}[t]
\centering
\small
\setlength{\tabcolsep}{2.4pt}
\begin{tabular}{@{}lcccc@{}}
\toprule
Run & FLOPs & H6 & MT-Bench & KL \\
\midrule
Source (Orion-Chat)            & --  & 58.93 & 7.49 & --    \\
\midrule
\multicolumn{5}{@{}l}{Normalizer initialization (trained with XKD)} \\
OPM ($\theta=\gamma$)          & 4.8 & 56.85 & 6.93 & 0.058 \\
Unit ($\theta=1$)              & 4.8 & 29.07 & 1.00 & 7.242 \\
Recentering                    & 4.8 & 58.02 & 7.17 & 0.054 \\
\midrule
\multicolumn{5}{@{}l}{Training signal (from the OPM initialization)} \\
UT, matched steps              & 3.6 & 56.77 & 2.80 & 0.248 \\
UT, matched steps, LR$\times$3 & 3.6 & 54.50 & 4.83 & 0.381 \\
UT, matched FLOPs              & 4.8 & 57.01 & 6.46 & 0.279 \\
Sequence-level KD              & 4.7 & 56.75 & 2.64 & 0.131 \\
\bottomrule
\end{tabular}
\caption{Controlled comparisons on the Chat variant. Runs use 1K steps except UT at matched FLOPs, which uses 1,333. LR$\times$3 triples the learning rate to $3{\times}10^{-5}$. FLOPs are in units of $10^{17}$ and include the teacher forward pass or, for sequence-level KD, generation. MT-Bench is the mean of three judging passes, and KL is the divergence from the source's next-token distribution on held-out text. Scores are comparable only within this table (Section~\ref{sec:experimental_settings}).}
\label{tab:controlled}
\end{table}

Table~\ref{tab:controlled} separates two choices that the 30K-step runs leave entangled, the normalizer initialization and the training signal. XKD here uses the default logit objective, and the evaluation protocol is that of Section~\ref{sec:experimental_settings}.

\paragraph{Normalizer initialization.} We distill from three initializations that differ only in how the normalizers are mapped. OPM copies $\gamma$. The unit initialization sets $\theta = 1$, the default for a newly created RMSNorm. Recentering additionally removes the mean from the residual-stream writers, which makes the mapping exact in $\gamma$ but modifies copied tensors (Section~\ref{sec:optimized_parameter_mapping}). Before any training, the KL divergence from the source is 3.70 for OPM, 4.04 for recentering, and 11.53 for the unit initialization. Random weights and the least-squares weights estimated from activation statistics start at 39.8 and 4.61. After 1K steps the unit initialization is still unusable, with an H6 score of 29.07 and the minimum MT-Bench score of 1.00, whereas OPM reaches 56.85 and $6.93 \pm 0.02$. Top-1 agreement with the source tells the same story, at 89.2\% for OPM and 8.3\% for the unit initialization. A budget that suffices from OPM therefore does not suffice from the default initialization. Recentering finishes somewhat ahead of OPM (58.02 and $7.17 \pm 0.03$), which makes it a reasonable alternative when modifying the copied weights is acceptable. OPM is the variant that leaves every copied tensor byte-identical.

\paragraph{Uptraining at matched computation.} At matched steps, UT uses $3.6{\times}10^{17}$ FLOPs against $4.8{\times}10^{17}$ for XKD and reaches an MT-Bench score of only 2.80. Tripling its learning rate raises the score to 4.83 but lowers H6 to 54.50 and moves the model further from the source. Extending UT to 1,333 steps at the default learning rate equalizes FLOPs and raises its score to $6.46 \pm 0.01$, still below the $6.93 \pm 0.02$ of XKD. H6 barely separates XKD from UT at the default learning rate (56.85 against 56.77 and 57.01). In all three UT runs the KL divergence from the source is four to six times that of XKD. At equal computation, XKD thus remains closer to the source and ahead on MT-Bench, although its margin of 0.47 points is smaller than the 30K-step margin in Table~\ref{tab:mt_bench}. Two learning rates do not amount to a full hyperparameter search for UT.

\paragraph{Sequence-level distillation.} In sequence-level KD~\citep{kim2016sequence}, the source samples continuations of the same prefixes and the student is trained on them with cross-entropy under the same optimizer and step count. H6 is indistinguishable from XKD (56.75 against 56.85), but MT-Bench falls to $2.64 \pm 0.07$ and the KL divergence rises to 0.131. At this budget, matching the source's full next-token distribution is far more effective than matching its samples.

These runs are short and use a single seed, so they establish orderings at 1K steps and not at convergence.

\begin{figure}[t]
  \centering
  \includegraphics[width=0.8\linewidth]{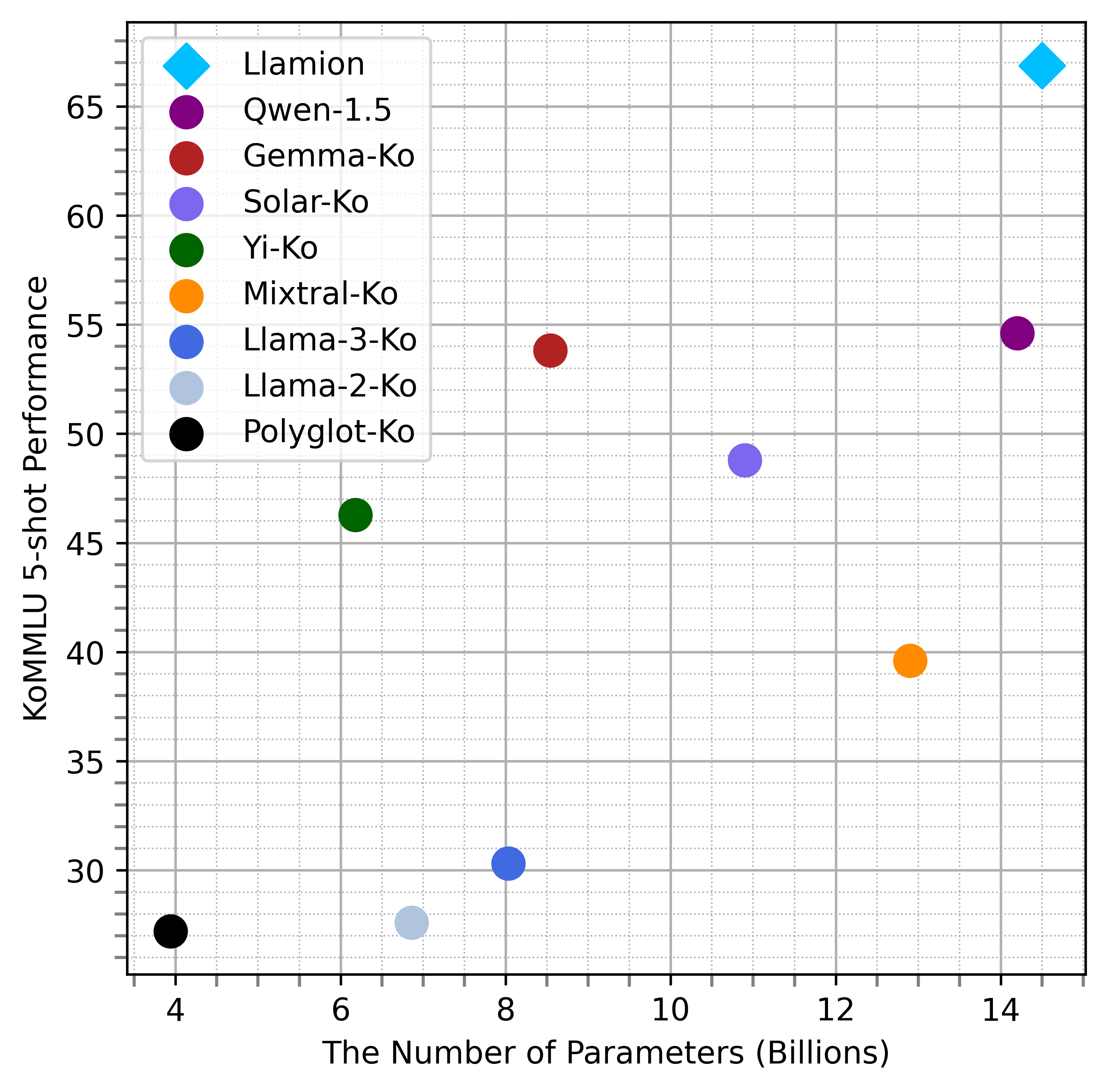}
  \caption{KoMMLU accuracy versus model size for representative open families on the leaderboard. The ${\sim}59.23\%$ next-best same-class entry is not among those shown.}
  \label{fig:kommlu}
\end{figure}

\begin{figure*}[t]
  \centering
  \includegraphics[width=0.48\linewidth]{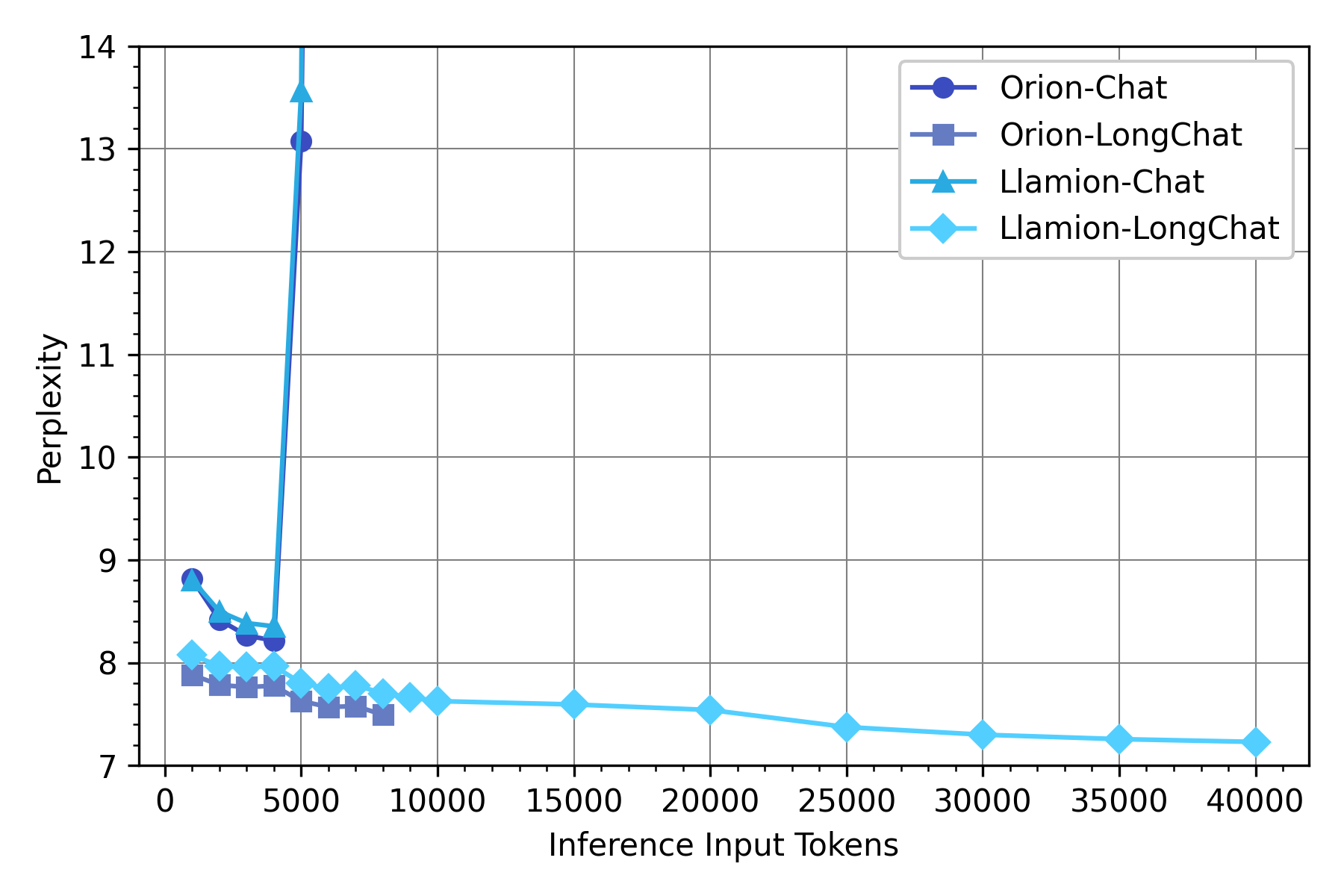}\hspace{0.03\linewidth}
  \includegraphics[width=0.48\linewidth]{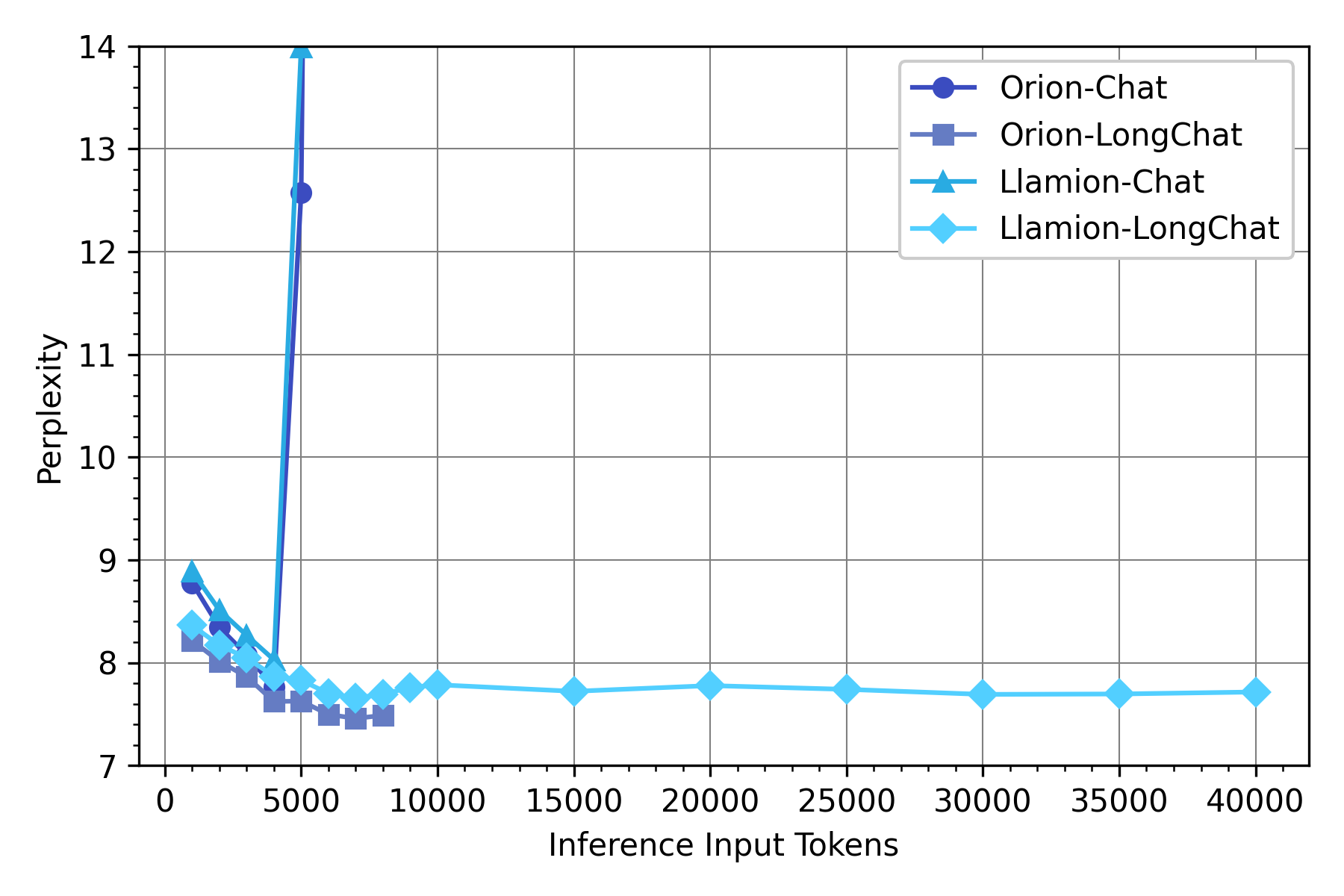}
  \caption{Perplexity versus input length on English Wikipedia (left) and Korean Wikipedia (right). Llamion-LongChat tracks Orion-LongChat closely in both languages up to 40K tokens, the longest length we evaluated, despite XKD using only 4K sequences. The standard Chat variants of both Orion and Llamion exhibit the expected PPL blow-up beyond ${\sim}5$K tokens, consistent with the fact that neither was adapted for long context.}
  \label{fig:wiki_ppl}
\end{figure*}

\subsection{Korean MMLU}
\label{sec:exp_kommlu}

The engineering motivation was a Korean-strong model admissible to \texttt{trust\_remote\_code\,=\,False} stacks, including the Open Ko LLM Leaderboard~\citep{park2024open}. We focus on KoMMLU, the leaderboard's Korean-translated MMLU subset. Llamion-Base's registered score is 66.87\%, exceeding the next-best same-class entry (${\sim}59.23\%$) by more than seven points at submission time (Figure~\ref{fig:kommlu}), far above the typical 1--2-point between-checkpoint variance. Orion itself has no registered score, because the leaderboard rejects custom-loader models. The conversion removes this restriction, so the score can be registered only for the converted model. The leaderboard score by itself establishes the converted model's absolute Korean capability. To measure preservation, we also evaluated Orion-Chat and Llamion-Chat locally on the Korean subset of MMMLU~\citep{openai2024mmmlu}, a separately translated Korean MMLU, with identical 5-shot inputs. They score 57.11 and 56.96. The difference of $-0.15$ points has a 95\% paired bootstrap interval of $[-0.59, +0.31]$, which contains zero. The fine-tuning evaluation in Appendix~\ref{sec:appendix_ft_korean} addresses contamination, using anonymized, redacted real user queries unlikely to appear in any pretraining corpus, and shows that the Korean capability persists under subsequent fine-tuning.

\section{Capability Preservation Beyond the Transfer Corpus}
\label{sec:analysis}

The MIRACL transfer corpus contains essentially no programming text and no multi-turn dialogue structure, and is capped at 4K-token sequences, yet the converted model retains all three capabilities. Two mechanisms can carry an out-of-corpus capability through conversion, and we interpret each case through them. Under \emph{initialization inheritance}, the parameters carrying a capability are copied and perturbed too little to destroy it. Under \emph{alignment maintenance}, XKD matches the teacher's outputs, so gradients never reward drift from teacher behavior, whereas uptraining fits its own corpus distribution and so pulls the model toward it. The evidence supports only the bounded claim that the conversion did not destroy these capabilities while uptraining with the same data and optimizer steps measurably eroded one. We do not claim behavior alignment on unseen inputs.

\paragraph{Multi-turn dialogue.}
Dialogue is the out-of-corpus capability we can quantify directly (Table~\ref{tab:mt_bench}). XKD preserves Orion-Chat's MT-Bench score within 0.1 points, while uptraining on the identical corpus and budget loses 1.4 points ($7.116 \to 5.709$). The comparison shares initialization, data, and step budget, which points to the training objective as the cause of uptraining's erosion, and the erosion persists when FLOPs are matched (Section~\ref{sec:exp_controlled}). In the three-pass re-evaluation, Orion-Chat and Llamion-Chat score $7.49 \pm 0.03$ and $7.45 \pm 0.02$, a paired difference of $-0.04$ with a 95\% bootstrap interval of $[-0.34, +0.26]$. How much of XKD's preservation is inheritance versus maintenance is not separated here, mirroring the long-context case.

\paragraph{Programming.}
On HumanEval~\citep{chen2021evaluating} and MBPP~\citep{austin2021program}, Llamion-Chat reaches pass@1 scores of 30.49 and 33.00 against 28.05 and 33.40 for Orion-Chat. The 95\% paired bootstrap intervals on the two differences, $[-1.83, +6.71]$ and $[-3.00, +2.20]$, both contain zero. Code generation is therefore preserved even though the transfer corpus contains essentially no code. Appendix~\ref{sec:appendix_examples} shows qualitative examples.

\paragraph{Long context.}
The Orion-LongChat source model supports inputs up to 200K tokens, while XKD used at most 4K. Figure~\ref{fig:wiki_ppl} reports perplexity on English and Korean Wikipedia~\citep{wikidump} versus input length, evaluated up to 40K tokens. Llamion-LongChat tracks Orion-LongChat in both languages across this range, $10\times$ the XKD length though short of the source's full window, and the parallel language curves rule out a transfer-corpus language artifact. We attribute this primarily to initialization inheritance. NPM preserves the RoPE base ($5{\times}10^{7}$ for LongChat) and all attention dimensions verbatim, large-base RoPE supports strong length extrapolation~\citep{liu2023scaling}, and XKD's role is to not damage what initialization inherited. The figure thus shows the conversion preserves long-context perplexity behavior, not that XKD could recover it had initialization destroyed it. On passkey retrieval~\citep{mohtashami2023landmark}, a downstream test of long-context use, Orion-LongChat and Llamion-LongChat both reach 100\% accuracy at 4K, 8K, 16K, 32K, and 64K tokens. The RoPE-reset ablation that would separate the two mechanisms remains unrun (Limitations).

\paragraph{Agreement with the source.}
Benchmark scores can agree while individual predictions differ, so we also compare the two models token by token. On held-out text, the KL divergence between the next-token distributions of Orion-Chat and Llamion-Chat is 0.038, and their top-1 predictions coincide at 91.4\% of positions. Their negative log-likelihoods on this text are nearly identical (2.246 and 2.251). The 1K-step XKD run of Table~\ref{tab:controlled} already reaches a KL divergence of 0.058 and 89.2\% top-1 agreement, which indicates that most of the alignment happens early in training.

\section{Discussion}
\label{sec:discussion}

\paragraph{Why alignment preserves more than uptraining.}
Uptraining fits the transfer corpus, so capabilities outside it receive no gradient signal that protects them. XKD instead matches the teacher's per-token outputs, so capabilities that the corpus does not exercise are disturbed only incidentally. This predicts that such capabilities are preserved, which is a weaker claim than alignment on arbitrary inputs.

\paragraph{When does an analytic OPM exist?}
The LN$\to$RMSNorm result instantiates a general recipe (identify a pretraining-induced input regularity, solve the population-level L2 alignment in closed form under it, and validate it against the least-squares optimum estimated from measured activations) that Appendix~\ref{sec:appendix_opm_extensions} applies to two further cases, a surrogate-level optimum for MHA$\to$GQA (mean-pooled KV projections) and a unit-level impossibility for bias-free GeLU$\to$SwiGLU. The pattern is that analytic OPM exists when the target's function class (nearly) contains the source's on the pretraining-induced regime, and distillation takes over where it does not.

\paragraph{Scope and compositionality with uptraining.}
KEPT requires a shared stacked-block, residual-stream structure and hidden dimension. Cross-width or cross-tokenizer transfer would need projection or embedding correspondences not investigated here. Where OPM lacks a closed-form rule, brief uptraining of the changed module plus XKD on the rest is a natural compromise.

\section{Conclusion}
\label{sec:conclusion}

We introduced KEPT, parameter-level mapping (NPM/OPM) plus equal-size frozen-teacher distillation (XKD), a low-compute recipe for architecture standardization, instantiated on Orion$\to$Llama as the released Llamion checkpoints. KEPT recovers the source's H6 and MT-Bench behavior from ${\sim}123$M tokens on one A100, attains class-leading Korean capability (66.87\% KoMMLU, a score its custom-loader source cannot register), beats uptraining at matched data and optimizer steps (most visibly on generative metrics), and preserves long-context perplexity behavior and dialogue, both absent from the transfer corpus. Controlled runs attribute the small budget to the analytic initialization, whose assumptions hold on three LayerNorm-based models, and show that the advantage over uptraining persists at matched FLOPs. If the recipe replicates on further source models, such conversions could be performed without pretraining-scale data or compute.

\section*{Limitations}

The full conversion is demonstrated on one source--target pair, and the 30K-step results come from single training runs. The controlled comparisons use 1K steps and a single seed, so they establish orderings at that budget only, and uptraining was tried at two learning rates, not tuned. Activation statistics on Falcon-7B and Pythia-12B support the assumptions behind OPM, but neither model was converted. KL-divergence logit distillation and the RoPE-reset ablation remain unrun. Long-context evaluation is limited to perplexity up to 40K tokens and passkey retrieval up to 64K, and the Korean fine-tuning suite is private. Absolute MT-Bench scores differ between our two evaluation passes, so they are comparable only within a table. Harder conversions are analyzed (Appendix~\ref{sec:appendix_opm_extensions}), not demonstrated.

\section*{Ethical Considerations}

KEPT preserves a source model's behavior as a whole, so its biases, unsafe behaviors, and memorized data carry over, and cheap conversion can prolong the life of weakly aligned models. Converted checkpoints should be treated as new releases, with safety and bias evaluations rerun rather than inherited. Llamion inherits Orion-14B's safety profile. The benefit is that multilingual capability which would otherwise be stranded is preserved at a fraction of uptraining's compute.

\section*{Acknowledgments}
This work was supported by the Institute of Information \& Communications Technology Planning \& Evaluation (IITP) grant funded by the Korean government (MSIT) (RS-2024-00397085, Leading Generative AI Human Resources Development).

\bibliography{custom}

\clearpage
\appendix

\section{Module-Level Mapping}
\label{sec:module_mapping}

\begin{table*}[t]
\centering
\small
\renewcommand{\arraystretch}{1.25}
\begin{tabular}{>{\raggedright\arraybackslash}p{0.15\linewidth}
                >{\raggedright\arraybackslash}p{0.27\linewidth}
                >{\raggedright\arraybackslash}p{0.30\linewidth}
                >{\raggedright\arraybackslash}p{0.16\linewidth}}
\toprule
\textbf{Module} & \textbf{Orion (source)} & \textbf{Llama (target)} & \textbf{Mapping rule} \\
\midrule
Token embedding    & $V \times d$                                 & $V \times d$                      & NPM (identity) \\
LM head            & $d \times V$                                 & $d \times V$                      & NPM (identity) \\
Attention proj.    & MHA, 40 heads                                & MHA, 40 heads                     & NPM (identity) \\
RoPE base          & Per-variant ($5 \times 10^{7}$ for LongChat) & Identical to source               & Preserved \\
Attention norm     & LayerNorm                                    & RMSNorm                           & \textbf{OPM} \\
MLP (gate/up/down) & SwiGLU                                       & SwiGLU                            & NPM (identity) \\
Pre-MLP norm       & LayerNorm                                    & RMSNorm                           & \textbf{OPM} \\
Final norm         & LayerNorm                                    & RMSNorm                           & \textbf{OPM} \\
Tokenizer          & Orion BPE, $V = 84{,}608$                    & Same vocab; Llama-3 template and special-token names & Vocab unchanged \\
\bottomrule
\end{tabular}
\caption{Per-module mapping between Orion (source) and Llama (target). All modules are mapped identity-wise except for normalization layers, which use OPM (Section~\ref{sec:optimized_parameter_mapping}). No feature concatenation or fusion is involved.}
\label{tab:module_mapping}
\end{table*}

Table~\ref{tab:module_mapping} consolidates the per-module rule, NPM (identity) for every module except the three normalizer site types (81 sites, two per layer plus the final norm), which use OPM. No module is structurally altered beyond the normalizer replacement.

\section{Proof of Optimized Parameter Mapping}
\label{sec:appendix_proof}

We show that, under the expected L2-output-alignment objective and the near-zero-mean activation assumptions below, the optimal RMSNorm weights $\theta^*$ for replacing a LayerNorm with weights $\gamma$ (and bias $\beta$) are equal to $\gamma$ up to negligible corrections, and the irreducible alignment residual is the discarded bias (exactly so in the zero-mean limit). We formulate the objective as the population-level expectation over the activation distribution, the appropriate formulation since $\theta$ is a single parameter vector shared across all inputs.

\paragraph{Setup.} Let $x = [x_1, \ldots, x_n]$ be drawn from the pre-norm activation distribution $\mathcal{D}$ at a given normalizer site, where $n$ is the hidden width. Define $\mu(x) = \tfrac{1}{n}\sum_{i} x_i$, $\sigma(x) = \sqrt{\tfrac{1}{n}\sum_{i}(x_i - \mu)^2}$, and $R(x) = \sqrt{\tfrac{1}{n}\sum_{i} x_i^2}$, so $R^2 = \sigma^2 + \mu^2$. LayerNorm outputs $Y_i = \tfrac{x_i - \mu}{\sigma}\gamma_i + \beta_i$; RMSNorm outputs $Y'_i = \tfrac{x_i}{R}\theta_i$.

\paragraph{Assumptions.}
\emph{(A1) Near-zero-mean regime:} $|\mu(x)| \le \varepsilon\,\sigma(x)$ \emph{almost surely} under $\mathcal{D}$ for some small $\varepsilon$, together with non-degeneracy: all expectations below are finite and $\mathbb{E}[x_i^2/R^2] > 0$ for every $i$. The almost-sure form allows replacing $\sigma$ by $R$ inside expectations directly (a high-probability bound would need additional tail control, though the bounded integrands make that straightforward). The regime is motivated by training with AdamW and non-trivial weight decay (Orion's setting, per its report), and the measurements below confirm it at the 99th percentile with $\varepsilon \approx 0.01$.
\emph{(A2) Averaged symmetry:} $\mathbb{E}[x_i/R] = 0$ and $\sum_{j \neq i} \mathbb{E}[x_i x_j / R^2] = 0$ for every $i$. Conditional sign symmetry of each coordinate given the rest is a sufficient idealization ($R$ is even in each coordinate, making both integrands odd), but is stronger than needed: (A2) tolerates strong pairwise correlations so long as their $R$-weighted signed average vanishes. We test (A2) through its consequence, namely that the least-squares optimum estimated from measured moments should be close to $\gamma$ (see the verification paragraph below). One consistency remark is that, since $\sum_j x_j = n\mu$ identically, (A2)'s second condition forces $\mathbb{E}[x_i \mu / R^2] = \mathbb{E}[x_i^2/R^2]/n > 0$, so (A1) and (A2) are jointly satisfiable only for $\varepsilon \gtrsim 1/n$; the exact $\mu \equiv 0$ limit lies outside their joint range, and the $O(\varepsilon^2)$ term below is at least $O(1/n^2)$, which the $1/n$ correction dominates (under (A2), the $(1-\tfrac{1}{n})$ factor is exact). The recentering construction of \citet{jiang2023prermsnorm} (Section~\ref{sec:optimized_parameter_mapping}) does not rely on these assumptions. With $\mu \equiv 0$ enforced pointwise, $\sigma = R$ and $\mathbb{E}[x_i\mu/R^2] = 0$, direct computation gives $\theta^* = \gamma$ exactly, with no $1/n$ correction, using only (A2)'s first condition.

\paragraph{Objective and solution.} Minimize the population objective
\[
J(\theta) = \mathbb{E}_{x \sim \mathcal{D}} \big\| Y - Y' \big\|_2^2 .
\]
$J$ is separable and strictly convex in each $\theta_i$; setting $\partial J / \partial \theta_i = 0$ gives
\[
\theta_i^{*}
= \gamma_i \frac{\mathbb{E}\!\left[\frac{x_i (x_i - \mu)}{\sigma R}\right]}{\mathbb{E}\!\left[\frac{x_i^{2}}{R^{2}}\right]}
\;+\;
\beta_i \frac{\mathbb{E}\!\left[\frac{x_i}{R}\right]}{\mathbb{E}\!\left[\frac{x_i^{2}}{R^{2}}\right]} .
\]
The $\beta$ term vanishes by (A2)'s first condition, regardless of the magnitude of $\beta$. The bias does not change the optimal weights and only contributes irreducible residual (below). For the $\gamma$ term, (A1) gives the pointwise bound $R/\sigma = \sqrt{1 + \mu^2/\sigma^2} \in [1, \sqrt{1+\varepsilon^2}]$ almost surely, so $\tfrac{x_i(x_i-\mu)}{\sigma R} = \big(1 + O(\varepsilon^2)\big)\tfrac{x_i(x_i-\mu)}{R^2}$ almost surely and the replacement is valid inside the expectation; then, expanding $\mu = \tfrac{1}{n}\sum_j x_j$,
\[
\mathbb{E}\!\left[\tfrac{x_i(x_i-\mu)}{R^2}\right]
= \big(1 - \tfrac{1}{n}\big)\,\mathbb{E}\!\left[\tfrac{x_i^2}{R^2}\right]
- \tfrac{1}{n}\!\sum_{j \neq i} \mathbb{E}\!\left[\tfrac{x_i x_j}{R^2}\right],
\]
where the cross-term sum vanishes by (A2)'s second condition. Hence
\[
\theta_i^{*} \;=\; \big(1 - \tfrac{1}{n}\big)\,\gamma_i \;+\; O(\varepsilon^2)
\;\approx\; \gamma_i ,
\]
with the $1/n$ correction equal to ${\sim}0.02\%$ at Orion's width $n = 5{,}120$, far below any other source of error. Setting $\theta \leftarrow \gamma$ is therefore optimal up to $O(1/n) + O(\varepsilon^2)$; wherever the paper says ``L2-optimal,'' it refers to this statement, absorbing the negligible corrections.

\paragraph{Irreducible residual.} At $\theta = \gamma$ and $\mu \equiv 0$ exactly, $Y_i - Y'_i = \beta_i$ for every input, so $J(\gamma) = \|\beta\|_2^2$: the alignment error that OPM cannot remove is the discarded additive bias, which RMSNorm's function class cannot represent. In the theoretical limit of a bias also driven to zero by weight decay, OPM alone would be exact; empirically it is not, as the mapping-only rows of Tables~\ref{tab:h6}--\ref{tab:mt_bench} show a large behavioral gap, indicating that the per-site residuals ($\|\beta\|_2^2$ at each of the $2 \times 40 + 1$ normalizer sites, plus the finite-$\mu$ corrections) compound substantially through 40 layers of depth. The two facts are consistent. (A1)--(A2) concern the weights that minimize the alignment error, a statement the measurements below support, while the size of the remaining error is governed by $\beta$ and the compounding. This residual is what the subsequent XKD stage absorbs by adjusting the surrounding (copied) modules (Section~\ref{sec:exp_h6}).

\paragraph{Empirical verification.} Table~\ref{tab:act_stats} reports pre-norm activation statistics at every normalizer site of three LayerNorm-based LLMs, measured on held-out text, with medians taken over sites. In all three models the mean is small relative to the standard deviation, with the 99th percentile of $|\mu|/\sigma$ at most 0.046, and the least-squares optimal RMSNorm weights computed from the measured moments lie within a few percent of $\gamma$. Orion is measured on text in the four transfer languages, and Falcon-7B and Pythia-12B on English text. In each model the alignment error at $\theta = \gamma$ is within about 2\% of $\|\beta\|_2^2$, while the error at the unit initialization is two orders of magnitude larger. The smallest observed $R^2$ on Orion is $8.4{\times}10^{4}$ times the stabilizing constant $\epsilon_0 = 10^{-5}$ that implementations add to the normalizer denominators, which justifies omitting $\epsilon_0$ from the setup above.

The deviation of the least-squares weights from $\gamma$ (1.9\% on Orion) exceeds the $1/n$ correction, so (A2) holds approximately and not exactly. This deviation offers nothing to exploit in practice. Initializing from the least-squares weights (denoted $\theta_{\mathrm{LS}}$ in Table~\ref{tab:act_stats}) gives a higher KL divergence from the source before training than $\theta = \gamma$ does (4.61 against 3.70, Section~\ref{sec:exp_controlled}), which is possible because the site-wise objective does not account for how errors compound across layers.

\begin{table}[t]
\centering
\small
\setlength{\tabcolsep}{3pt}
\begin{tabular}{@{}lcccccc@{}}
\toprule
Model & Sites & $|\mu|/\sigma$ & $\theta_{\mathrm{LS}}$ vs.\ $\gamma$ & $J(\gamma)$ & $\|\beta\|_2^2$ & $J(1)$ \\
\midrule
Orion-14B  & 81 & 0.010 & 1.9\% & 12.5 & 12.2 & 2,126 \\
Falcon-7B  & 33 & 0.046 & 3.4\% & 13.1 & 12.8 & 2,042 \\
Pythia-12B & 73 & 0.012 & 2.2\% & 19.8 & 19.4 & 2,416 \\
\bottomrule
\end{tabular}
\caption{Pre-norm activation statistics (medians over LayerNorm sites). $|\mu|/\sigma$ is its 99th percentile, $\theta_{\mathrm{LS}}$ vs.\ $\gamma$ is the relative distance of the least-squares optimal weights from $\gamma$, and $J(\gamma)$ and $J(1)$ are the alignment errors at $\theta=\gamma$ and at the unit initialization.}
\label{tab:act_stats}
\end{table}

We also trained the normalizer weights alone with all other parameters frozen. From random weights, 100 steps reduce the KL divergence from the source only from 38.72 to 35.28. From $\theta = \gamma$, the same 100 steps reduce it from 4.03 to 0.27. (This probe uses a smaller held-out sample than Table~\ref{tab:controlled}, hence 4.03 instead of 3.70 at the start.) The analytic initialization is therefore a far better starting point than learning the normalizers from scratch, and the normalizer weights alone can absorb a large part of the residual.

\section{When Does an Analytic OPM Exist? Extensions Beyond Normalization}
\label{sec:appendix_opm_extensions}

This appendix develops the mathematical account of when the OPM approach extends to module substitutions other than LN$\to$RMSNorm. None of the extensions below are demonstrated empirically in this paper; they are derivations and impossibility arguments that delineate the boundary of the analytic recipe.

\paragraph{The general recipe.} An analytic OPM for a module substitution $f_{src} \to f_{tgt}$ consists of three steps. \emph{(i) Identify a pretraining-induced regularity} of the module's input distribution $\mathcal{D}$, a property that training itself creates rather than one assumed of the data (for LN$\to$RMSNorm, near-zero-mean activations under weight decay). \emph{(ii) Solve the population alignment} $\min_{\phi} \mathbb{E}_{x\sim\mathcal{D}} \| f_{src}(x) - f_{tgt}(x; \phi) \|^2$ in closed form under that regularity. \emph{(iii) Validate empirically}: compare the analytic solution with the least-squares optimum estimated from measured activations, and check that training the new module alone from the analytic start reduces the divergence from the source (Appendix~\ref{sec:appendix_proof}, last paragraph). An analytic OPM exists precisely when step (ii) has a closed-form solution whose residual is small on $\mathcal{D}$, equivalently when the target module's function class (nearly) contains the source module's function on the activation regime pretraining induces, even if not on all of $\mathbb{R}^n$.

\subsection{A Positive Result: MHA$\to$GQA Mean-Pooling Is L2-Optimal at the Projection Level}
\label{sec:appendix_gqa}

Grouped-query attention~\citep{ainslie2023gqa} merges the per-head projections $W_1, \ldots, W_h \in \mathbb{R}^{d_k \times d}$ of each group of $h$ key (resp.\ value) heads into a single shared projection $W$, where $d$ is the hidden width, $d_k$ the per-head dimension, and $x \in \mathbb{R}^{d}$ the input to the projection. Take as surrogate objective the summed alignment error of the projected keys over the group:
\begin{align*}
J(W) &= \sum_{i=1}^{h} \mathbb{E}_{x}\, \big\| W x - W_i x \big\|_2^2 \\
&= \sum_{i=1}^{h} \operatorname{tr}\!\big[ (W - W_i)\, \Sigma\, (W - W_i)^{\!\top} \big],
\end{align*}
where $\Sigma = \mathbb{E}[x x^{\top}]$ is the input second-moment matrix. Setting the gradient $\partial J/\partial W = 2 \sum_i (W - W_i)\Sigma$ to zero gives $\big(hW - \sum_i W_i\big)\Sigma = 0$, so for any non-singular $\Sigma$,
\[
W^{*} = \frac{1}{h} \sum_{i=1}^{h} W_i ,
\]
i.e.\ \emph{mean-pooling of the per-head projections is the unique L2-optimal shared projection, independently of the input's second-moment matrix}. The same argument applies verbatim to value projections. This recovers, as the analytic optimum of step (ii), exactly the mean-pooling initialization that \citet{ainslie2023gqa} adopted heuristically and found empirically superior to selecting a single head. Two caveats keep this a surrogate-level result. The objective aligns pre-attention projections rather than post-softmax attention outputs (a small key perturbation is scaled by query magnitude before the softmax), and the query projections are untouched. A sharper characterization strengthens both. At the pre-softmax logit level, writing $W_{q,i}$ for the query projection of head $i$ and $\Sigma_q$, $\Sigma_k$ for the second-moment matrices of the query and key inputs, treating query and key inputs as decoupled, and assuming $\Sigma_k$ and $\sum_i A_i$ non-singular (the latter holds whenever no key-space direction is ignored by every head in the group), minimizing the summed logit-alignment error yields the weighted optimum $W^{*} = \big(\sum_i A_i\big)^{-1} \sum_i A_i W_i$ with $A_i = W_{q,i}\,\Sigma_q\,W_{q,i}^{\top}$ the $i$-th head's query second-moment matrix; mean-pooling is the special case of equal $A_i$, i.e.\ the data-free choice available without estimating query statistics. The key and value cases are also asymmetric. Holding attention weights fixed, the head output is a convex combination of projected values, so value-projection error bounds the output error with constant one, whereas key error enters through the logits. The XKD stage is designed to absorb all of these residuals, so the natural KEPT pipeline for MHA$\to$GQA is NPM elsewhere, mean-pooling (or the weighted variant) for merged KV heads, then XKD. This is the most direct next case study for the framework and is not a demonstrated result.

\subsection{A Negative Result: No Exact Unit-Level GeLU$\to$SwiGLU Mapping}
\label{sec:appendix_swiglu}

A GeLU MLP computes $W_2\,\mathrm{GeLU}(W_1 x)$, where $W_1$ and $W_2$ are its input and output projections, $\mathrm{GeLU}(a) = a\,\Phi(a)$, and $\Phi$ is the standard normal distribution function. Consider one hidden unit with input weights $w_1$ (a row of $W_1$) along a ray $x = tv$ with direction $v$ and scale $t > 0$. The unit grows \emph{linearly} in $t$, since $\mathrm{GeLU}(\hat{a}t) \sim \hat{a}t$ as $t \to +\infty$ for $\hat a = w_1\!\cdot\!v > 0$. A bias-free SwiGLU MLP, the Llama-template convention, computes $W_2\big(\mathrm{SiLU}(W_g x) \odot (W_u x)\big)$ with gate and up projections $W_g$ and $W_u$, where $\mathrm{SiLU}(g) = g\,s(g)$ and $s$ is the logistic sigmoid. Each hidden coordinate is $\mathrm{SiLU}(g)\,u$ with $g = w_g\!\cdot\!x$ and $u = w_u\!\cdot\!x$ both linear and bias-free ($w_g$ and $w_u$ are rows of $W_g$ and $W_u$). Along the same ray, with $\hat g = w_g\!\cdot\!v$ and $\hat u = w_u\!\cdot\!v$, it grows \emph{quadratically}, as $\mathrm{SiLU}(\hat{g}t)\,\hat{u}t \sim \hat{g}\hat{u}\,t^2$. Exact equality for all $x$ therefore forces the leading coefficient $\hat{g}(v)\,\hat{u}(v)$ to vanish on the open half-space $\{\hat g > 0\}$, hence everywhere, since a polynomial vanishing on an open set vanishes identically; a product of two linear forms is identically zero only if one form is, i.e.\ $w_g = 0$ or $w_u = 0$, either of which makes the unit identically zero. \emph{No non-degenerate $(w_g, w_u)$ reproduces a GeLU unit exactly.} We scope the formal claim to the unit level, since ruling out exact emulation of an entire GeLU MLP via cross-unit cancellation through $W_2$ would additionally require showing that every sign-cone's quadratic ray-asymptote vanishes and that the residual tails (sigmoid-type $e^{-ct}$ on the SwiGLU side versus Gaussian-type $e^{-ct^2/2}$ on the GeLU side) cannot coincide; we do not develop that longer argument here. The practical conclusion is unaffected, as no constructive closed-form mapping is available. The obstruction is architectural and lies in the bias-free convention. If the up-projection carried a bias, denoted $b_u$, setting $w_u = 0$, $b_u = 1/1.702$, $w_g = 1.702\,w_1$ would reproduce $a\,s(1.702a)$ with $a = w_1\!\cdot\!x$, the standard sigmoid approximation of GeLU accurate to ${\sim}10^{-2}$, yielding a usable approximate OPM. The Llama template's bias-free MLP rules out this route. Within the template, the mismatch is functional, not statistical, so no pretraining-induced input regularity (step i) can repair it and no closed-form step (ii) exists; the least-squares projection onto the SwiGLU class has no closed form and amounts to module-wise distillation, so the recipe prescribes a brief module-level training pass followed by XKD (Section~\ref{sec:discussion}). The common width-convention change (SwiGLU layers typically use $\tfrac{2}{3}$ of the GeLU MLP's intermediate width) compounds this with a dimension mismatch.

\subsection{Taxonomy of Module Substitutions}
\label{sec:appendix_taxonomy}

\begin{table}[t]
\centering
\small
\setlength{\tabcolsep}{3pt}
\renewcommand{\arraystretch}{1.2}
\begin{tabular}{@{}p{0.30\linewidth}p{0.32\linewidth}p{0.28\linewidth}@{}}
\toprule
\textbf{Substitution} & \textbf{Analytic OPM} & \textbf{Status} \\
\midrule
LN$\to$RMSNorm & $\theta = \gamma$ (App.~\ref{sec:appendix_proof}) & Derived; demonstrated (Section~\ref{sec:exp_h6}) \\
MHA$\to$GQA & mean-pooled KV projections (App.~\ref{sec:appendix_gqa}) & Derived (surrogate level); not demonstrated \\
GeLU$\to$SwiGLU & none at unit level (App.~\ref{sec:appendix_swiglu}) & Impossibility argument; fall back to distillation \\
Learned PE$\to$RoPE & none (no parameter correspondence) & Distillation-heavy \\
Depth insertion & identity-initialized blocks & Prior work~\citep{wu2024llamapro} \\
Width change & requires projection machinery & Future work \\
Cross-tokenizer & requires embedding correspondence & Future work \\
\bottomrule
\end{tabular}
\caption{Where the analytic-OPM recipe applies. ``Demonstrated'' means empirically exercised in this paper's case study; derivations without demonstration are labeled as such.}
\label{tab:opm_taxonomy}
\end{table}

Table~\ref{tab:opm_taxonomy} summarizes the resulting classification. Closed-form OPM exists where the target class contains (LN$\to$RMSNorm, under the induced regime) or linearly spans (GQA merging) the source computation, fails where the classes are functionally disjoint (GeLU$\to$SwiGLU), and is inapplicable where no parameter correspondence exists at all (positional-encoding replacement, tokenizer change). Distillation absorbs progressively more of the work as one moves down the table, which is also a statement about expected cost; the further a planned conversion sits from the top, the less of KEPT's efficiency argument should be assumed to carry over.

\section{Fine-tuning on Korean Generation}
\label{sec:appendix_ft_korean}

We complement the few-shot KoMMLU evaluation with a fine-tuning evaluation on four Korean generation tasks (Table~\ref{tab:four_tasks}), namely Closed-book QA, Title Generation, Document Summarization, and Open-book QA. The queries are anonymized real user queries from a Korean search system, redacted to remove PII; the gold answers are produced by GPT-4 with sampled manual correction (correction rate 5--8\%).

\begin{table}[t]
\centering
\small
\setlength{\tabcolsep}{3.5pt}
\begin{tabular}{@{}lcccc@{}}
\toprule

\multirow{2}{*}{\raisebox{-0.4ex}[0pt][0pt]{Task}} & \multicolumn{2}{c}{Inputs} & \multicolumn{2}{c}{Data Size}\\

\noalign{\vskip 2pt}
\cline{2-3}
\cline{4-5}
\noalign{\vskip 2pt}

     & Query & Doc(s) & \# Train & \# Test \\
\midrule

Closed-book QA & 1 & 0 & 10,229 & 600 \\
Title Generation & 0 & 1 & 21,776 & 200\\
Doc. Summarization & 0 & 2+ & 14,037 & 200\\
Open-book QA & 1 & 2+ & 25,052 & 300\\

\bottomrule
\end{tabular}
\caption{
    Input types and data statistics of the Korean generation tasks to measure fine-tuning performance
}
\label{tab:four_tasks}
\end{table}





\begin{table}[t]
\centering
\small
\begin{tabular}{lcc}

\toprule
\multicolumn{1}{c}{Model} & \# of Total   & \# of Korean  \\
\midrule
Polyglot-Ko 13B           & 167B    & 167B         \\
Llama-2 13B               & 2.0T  & 1B           \\
Mistral 7B                & Unknown & Unknown      \\
Gemma 7B                & 6.0T & Unknown      \\
\midrule
Orion 14B                 & 2.5T  & \textit{50B} \\
\bottomrule
\end{tabular}
\caption{
    Comparison of pretrained models showing the total number of pretraining tokens and the number of those tokens in Korean. \textit{Italic} values are estimates.
}
\label{tab:num_pretrained_tokens}
\end{table}

Fine-tuning uses maximum length 2{,}048, batch size 32, learning rate $2{\times}10^{-5}$ with cosine annealing and 1K warmup steps, and up to 70K steps. Responses are scored on a 0--10 scale by \texttt{gpt-4-0613} with single-answer grading and averaged within each task. Baselines are GPT-3.5 and GPT-4 via API, Meta's Llama-2-13B-Chat~\citep{touvron2023llama}, Google's Gemma-7B-IT~\citep{team2024gemma}, EleutherAI's Polyglot-Ko-13B~\citep{ko2023technical}, MistralAI's Mistral-7B-Instruct~\citep{jiang2023mistral}, Yanolja's EEVE-Korean-Instruct-10.8B~\citep{kim2024efficient}, and Upstage's SOLAR-10.7B-Instruct~\citep{kim2023solar}. Table~\ref{tab:num_pretrained_tokens} compares the pretraining-token composition of Orion with that of several of these baselines.

\begin{figure*}[t]
  \centering
  \includegraphics[width=0.95\linewidth]{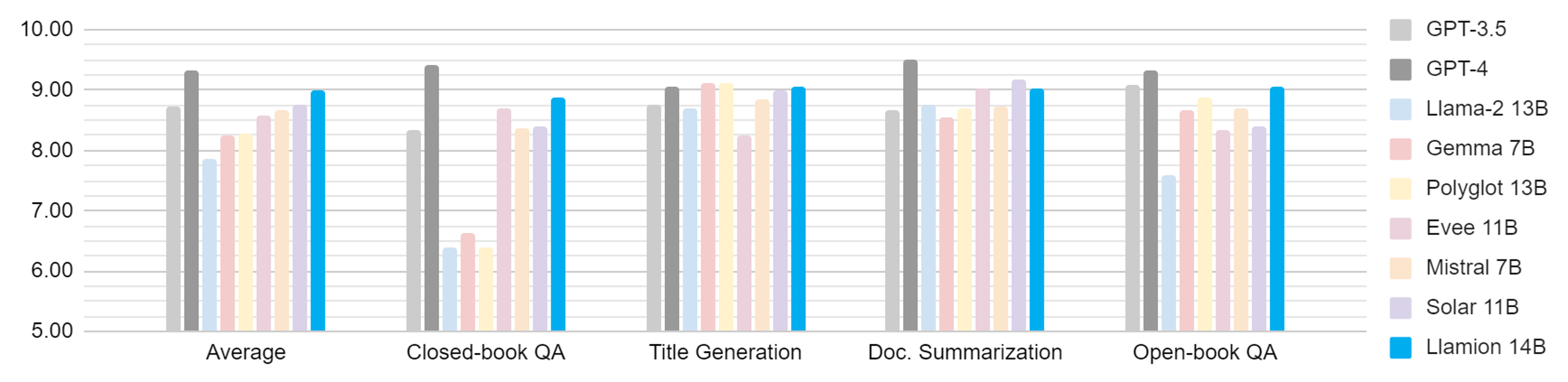}
  \caption{Fine-tuning performance on four Korean generation tasks (GPT-4-graded G-Eval). Fine-tuned Llamion-Chat outperforms GPT-3.5 on the four-task average and exceeds all open baselines, including those trained primarily on Korean.}
  \label{fig:ft_korean_four_tasks}
\end{figure*}

Figure~\ref{fig:ft_korean_four_tasks} shows that fine-tuned Llamion-Chat outperforms GPT-3.5 on the four-task average and exceeds all open baselines, including those trained primarily on Korean. On Closed-book QA in particular, the weakest baselines (Llama-2, Gemma, Polyglot) exhibit substantial hallucination; for the predominantly non-Korean Llama-2 and Gemma, this indicates that Korean-language factuality requires substantially more Korean pretraining data than they received. The evaluation serves two purposes. It confirms that the few-shot KoMMLU advantage transfers to a generative setting where contamination is implausible (queries are anonymized real user queries, not drawn from any pre-existing benchmark), and it shows that Llamion's Korean capability persists under subsequent fine-tuning, i.e.\ that KEPT preserves the underlying Korean-language competence in a form that fine-tuning can build on.

\section{Qualitative Examples of Out-of-Corpus Capability Preservation}
\label{sec:appendix_examples}

\begin{figure*}[t]
  \centering
  \includegraphics[width=0.9\textwidth]{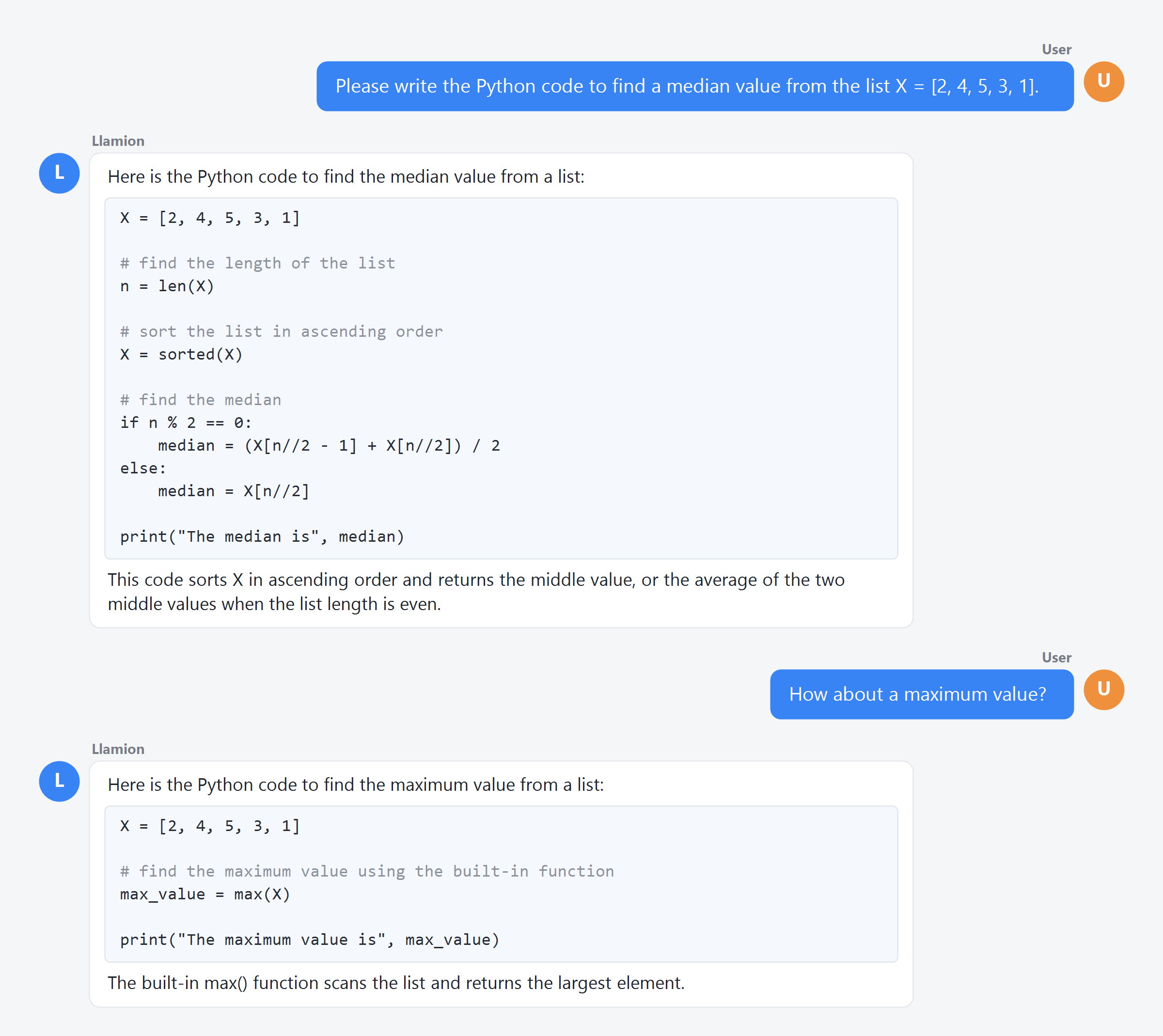}
  \caption{Representative interactions with Llamion-14B-Chat, showing coherent multi-turn dialogue and Python programs despite neither capability being represented in the XKD training corpus. Illustrative examples, not a benchmark evaluation.}
  \label{fig:python_llamion}
\end{figure*}

Capabilities not covered by the XKD training corpus, namely multi-turn dialogue and Python programming, are observed to survive the architectural transition. Figure~\ref{fig:python_llamion} shows representative interactions with Llamion-14B-Chat. These examples are illustrative, and the quantitative evidence is in Section~\ref{sec:analysis}.

\section{Reproducibility Details}
\label{sec:appendix_reproducibility}

This appendix consolidates the implementation details required to reproduce KEPT end-to-end on the Orion$\to$Llama case.

\paragraph{Hardware and memory.} A single NVIDIA A100 80GB GPU, with bf16 throughout training and evaluation. Wall-clock time for the full XKD run is approximately four days. All student parameters are updated. Two 14B models do not fit on one 80GB device together with optimizer state, so training uses DeepSpeed ZeRO stage 2~\citep{rajbhandari2020zero} with the optimizer offloaded to the CPU~\citep{ren2021zerooffload} and gradient checkpointing. The GPU holds the bf16 student (29 GB), the frozen bf16 teacher (29 GB), whose logits are computed on the fly, and activations. The fp32 master weights, gradients, and AdamW states (about 230 GB) stay in host memory. Measured peak GPU memory is 66.9 GB for XKD against 35.6 GB for uptraining, which keeps no teacher, and peak host memory is 239.5 GB.

\paragraph{Computation.} We count floating-point operations (FLOPs) with the usual approximation of $2P$ per token for a forward pass and $4P$ for a backward pass, where $P \approx 1.45{\times}10^{10}$ is the parameter count. An uptraining step over $T$ tokens then costs $6PT$, and an XKD step costs $8PT$ because of the additional teacher forward pass, a ratio of $1.33$. With $T = 4{,}096$ tokens per step, 1K steps cost $3.6{\times}10^{17}$ FLOPs for uptraining and $4.8{\times}10^{17}$ for XKD, and the full 30K-step runs cost about $1.1{\times}10^{19}$ and $1.4{\times}10^{19}$. The comparisons in Section~\ref{sec:exp_h6} match data and optimizer steps, so XKD uses about one third more computation than uptraining. In the controlled comparisons of Section~\ref{sec:exp_controlled}, uptraining at matched FLOPs runs for 1,333 steps, and sequence-level KD costs $4.7{\times}10^{17}$ FLOPs for 1K steps including the generation of its training sequences.

\paragraph{Source model and architectural target.} Source: Orion-14B (\texttt{OrionStarAI/Orion-14B-Base}, \texttt{Orion-14B-Chat}, \texttt{Orion-14B-LongChat}); architectural target: the Llama-family template (decoder-only, RoPE, RMSNorm, SwiGLU). The Llamion configuration matches Orion's dimensions exactly ($V=84{,}608$, $d=5{,}120$, MLP intermediate $15{,}360$, $40$ heads of dim $128$, $40$ layers, full multi-head attention, RoPE base $5{\times}10^{7}$ for LongChat).

\paragraph{Parameter mapping.} NPM is a direct tensor copy for token embedding, LM head, all attention projections, all MLP projections, and tokenizer (per Table~\ref{tab:module_mapping}). OPM for the three normalizer site types (81 sites in total) is $\theta \leftarrow \gamma$, $\beta$ discarded; no normalizer training is required (Appendix~\ref{sec:appendix_proof}). String-level changes to the tokenizer are confined to the chat template (Llama-3 format) and the assignment of Llama-3 special-token surface forms to existing special/reserved token IDs; the embedding matrix is unchanged. Concretely, in the released tokenizer configuration, \texttt{<|pad|>} occupies ID 0, \texttt{<|begin\_of\_text|>} and \texttt{<|end\_of\_text|>} occupy the source BOS/EOS IDs (1 and 2), and \texttt{<|eot\_id|>}, \texttt{<|start\_header\_id|>}, \texttt{<|end\_header\_id|>} occupy the low-index special-token IDs 5--7.

\paragraph{XKD training.} Corpus: MIRACL~\citep{10.1162/tacl_a_00595} restricted to Korean/English/Chinese/Japanese subsets, balanced to ${\sim}123$M tokens; sequences concatenated and packed to max length $4{,}096$~\citep{liu2019roberta}. Per-device batch size $1$ (the per-token logit is the effective sample). Optimizer: AdamW, learning rate $1{\times}10^{-5}$, cosine annealing, $100$ warmup steps, $30{,}000$ total steps. Learning rates above $3{\times}10^{-5}$ produced training instability, and we recommend matching the learning rate used during the source model's pretraining. Loss: either $\mathcal{L}_{h}$ (Eq.~\ref{eq:loss_hidden}) or $\mathcal{L}_{z}$ (Eq.~\ref{eq:loss_logits}), with the logit variant as our default. The teacher is frozen.

\paragraph{Evaluation.} H6: LM Evaluation Harness~\citep{eval-harness} with Open LLM Leaderboard~\citep{open-llm-leaderboard} few-shot configurations (ARC 25-shot, HellaSwag 10-shot, MMLU/Winogrande/GSM8K 5-shot, TruthfulQA 0-shot). MT-Bench: FastChat with the \texttt{gpt-4-0613} judge, single-answer grading, single run. KoMMLU: Open Ko LLM Leaderboard~\citep{park2024open}, registered scores from the public submission. Fine-tuning (Appendix~\ref{sec:appendix_ft_korean}): max length $2{,}048$, batch size $32$, learning rate $2{\times}10^{-5}$ cosine, $1{,}000$ warmup steps, up to $70{,}000$ steps, G-Eval scoring $0$--$10$ via \texttt{gpt-4-0613} averaged within each task. Controlled comparisons (Section~\ref{sec:exp_controlled}): 1K steps of 4{,}096 tokens (4.1M tokens) on the same stream, MT-Bench judged three times by \texttt{gpt-4-0613}, and KL divergence and top-1 agreement with the source computed on held-out text. Preservation benchmarks (Section~\ref{sec:analysis}) use LM Evaluation Harness with identical inputs for both models: the Korean subset of MMMLU with 5 shots, HumanEval with 0 shots, and MBPP with 3 shots. Paired bootstrap intervals use 5{,}000 resamples over items, or over questions for MT-Bench. Activation statistics (Table~\ref{tab:act_stats}) are computed over all LayerNorm modules of each model.

\paragraph{Released artifacts.} The Llamion-Base, Llamion-Chat, and Llamion-LongChat checkpoints are released on the Hugging Face Hub as the \texttt{vaiv/GeM2-Llamion} collection\footnote{\url{https://huggingface.co/collections/vaiv/gem2-llamion}}.
They load with \texttt{trust\_remote\_code=False} in the Transformers library. The KEPT training script, parameter-mapping utility, and evaluation harness configurations are released alongside the manuscript. The Korean fine-tuning evaluation in Appendix~\ref{sec:appendix_ft_korean} uses an internal user-query corpus that is not released for privacy reasons. The H6, MT-Bench, and KoMMLU evaluations are fully reproducible from public assets.

\end{document}